\documentclass[lettersize,journal]{IEEEtran}
\usepackage{amsmath,amsfonts}
\usepackage{algorithmic}
\usepackage{algorithm}
\usepackage{array}
\usepackage[caption=false,font=normalsize,labelfont=sf,textfont=sf]{subfig}
\usepackage{textcomp}
\usepackage{stfloats}
\usepackage{url}
\usepackage{verbatim}
\usepackage{graphicx}
\usepackage{amsthm}
\usepackage{orcidlink}
\newtheorem{definition}{Definition}
\DeclareUnicodeCharacter{2212}{$-$}
\graphicspath{{./figs/}}
\usepackage{cite}

\begin{document}

\title{Behavioral Safety Assessment towards Large-scale Deployment of Autonomous Vehicles, Part I: Methodology}

\author{Henry X. Liu\textsuperscript{\#}\,\orcidlink{0000-0002-3685-9920},~\IEEEmembership{Fellow,~IEEE,}
        Tinghan Wang\textsuperscript{\#}\,\orcidlink{0000-0002-5660-8889},
        Xintao Yan\textsuperscript{\#}\,\orcidlink{0000-0002-0569-5628},
        Haowei Sun\textsuperscript{\#}\,\orcidlink{0000-0002-0232-117X},
        Zhijie Qiao,
        Kenneth Boyd,
        Shuo Feng\,\orcidlink{0000-0002-2117-4427},
        Greg Stevens,
        Greg McGuire
\thanks{Henry X. Liu (e-mail: henryliu@umich.edu), Kenneth Boyd (e-mail: kjboyd@umich.edu), Greg Stevens (e-mail: gregstvn@umich.edu), and Greg McGuire (e-mail: mcguireg@umich.edu) are with the University of Michigan Transportation Research Institute, Ann Arbor, MI 48109 USA.\par
Henry X. Liu, Tinghan Wang (e-mail: tinghanw@umich.edu), Zhijie Qiao (e-mail: zhijieq@umich.edu) are with the Department of Civil and Environmental Engineering, University of Michigan, Ann Arbor, MI 48109 USA.\par
Xintao Yan (e-mail: xintaoy@hku.hk) is with the Department of Civil Engineering, The University of Hong Kong, Hong Kong 999077, China.\par
Haowei Sun (e-mail: howie@inchor.ai) is with Laplace Intelligence, Ann Arbor, MI 48109 USA.\par
Shuo Feng (e-mail: fshuo@tsinghua.edu.cn) is with the Department of Automation, Tsinghua University, Beijing 100084, China.\par
Corresponding author: Henry X. Liu (e-mail: henryliu@umich.edu).\par
\# Equal Contribution.}%
}

\maketitle

\begin{abstract}
Autonomous vehicles (AVs) have significantly advanced in real-world deployment in recent years, yet safety continues to be a critical barrier to widespread adoption. Traditional functional safety approaches, which primarily verify the reliability, robustness, and adequacy of AV hardware and software systems from a vehicle-centric perspective, do not sufficiently address the AV's broader interactions and behavioral impact on the surrounding traffic environment. To overcome this limitation, we propose a paradigm shift toward behavioral safety, a comprehensive approach focused on evaluating AV responses and interactions within the traffic environment. To systematically assess behavioral safety, we introduce a third-party AV safety assessment framework comprising two complementary evaluation components: the Behavioral Competency Test and the Driving Intelligence Test. The Behavioral Competency Test evaluates the AV's reactive behaviors under controlled scenarios, ensuring basic behavioral competency. In contrast, the Driving Intelligence Test assesses the AV's interactive behaviors within naturalistic traffic conditions, quantifying the frequency of safety-critical events to deliver statistically meaningful safety metrics before large-scale deployment. In Part II of this study~\cite{Liu2026partII}, an open-source Level 4 Automated Driving System (ADS) is tested to demonstrate the effectiveness of the proposed method.
\end{abstract}


\section{Introduction}
\label{sec: introduction}
\IEEEPARstart{A}{utonomous} vehicle (AV) technology is transforming modern transportation, promising increased safety, efficiency, and accessibility. In recent years, AV deployments have expanded globally. Waymo, for example, opened its robotaxi services to the public in 10 U.S. cities as of early 2026, including Phoenix, the San Francisco Bay Area, Los Angeles, Atlanta, Austin, and Miami, and plans to launch in further cities such as London later in 2026~\cite{korosec2026waymo}.
Similarly, Baidu's Apollo Go has launched operations in Wuhan and other Chinese cities~\cite{zhang2024baidu}.
However, these deployments have sparked widespread debate, particularly regarding AV safety~\cite{lu2023san}.
High-profile incidents, such as when a Cruise robotaxi in 2023 hit and dragged a pedestrian about 20 feet in San Francisco~\cite{thadani2023cruise}, have intensified public scrutiny and eroded trust in AV technology.
A recent survey shows that only 13\% of U.S. drivers trust riding in AVs~\cite{moye2025aaa}, highlighting the public's skepticism and the pressing need to build confidence.
Public acceptance and consumer confidence depend heavily on perceived safety~\cite{jing2020determinants}, making unbiased, rigorous, transparent, and publicly accessible safety evaluation programs essential.
Furthermore, systematic AV safety assessment frameworks are crucial for regulators to establish effective safety standards and facilitate the seamless integration of AVs into existing transportation systems.

\IEEEpubidadjcol

Despite the critical need, a systematic AV testing program, particularly one based on third-party evaluations, remains absent.
While well-established testing programs exist for conventional human-driven vehicles~\cite{nhtsa_ncap,euroncap,iihs_ratings,cncap,iso26262,iso21448,iso8800,iso34502,ul4600}, they are inadequate for Level 4--5 AVs~\cite{sae2021j3016}, as they adopt a vehicle-centric perspective focused on passive and active safety features or system-level malfunctions, rather than evaluating an AV's holistic driving behavior in complex traffic environments.
While essential, a vehicle-centric approach is insufficient for evaluating AV safety in large-scale real-world deployments for two key reasons.
First, large-scale deployment requires evaluating AVs' safety impact on the surrounding traffic environment, including not only their ability to mitigate or avoid hazards but also their capacity to avoid creating new safety hazards.
This necessitates an environment-based approach that evaluates how AVs react to and interact with dynamic agents in stochastic, diverse, and complex traffic situations, an aspect Functional Safety (FuSa) and Safety of the Intended Functionality (SOTIF) do not address.
Second, AV testing must shift from deterministic assessment to probabilistic evaluation, as modern AVs rely heavily on deep learning models that exhibit inherently non-deterministic behavior.
In other words, safety evaluation based solely on predetermined scenarios is insufficient, and unbiased estimation of the overall crash rate is necessary to justify the net societal benefits of AVs and to prevent damage to consumer trust from singular and unreasonable risks.
It also helps answer the fundamental question, ``How safe is safe enough?''

\begin{figure*}[t]
  \centering
  \includegraphics[width=\textwidth]{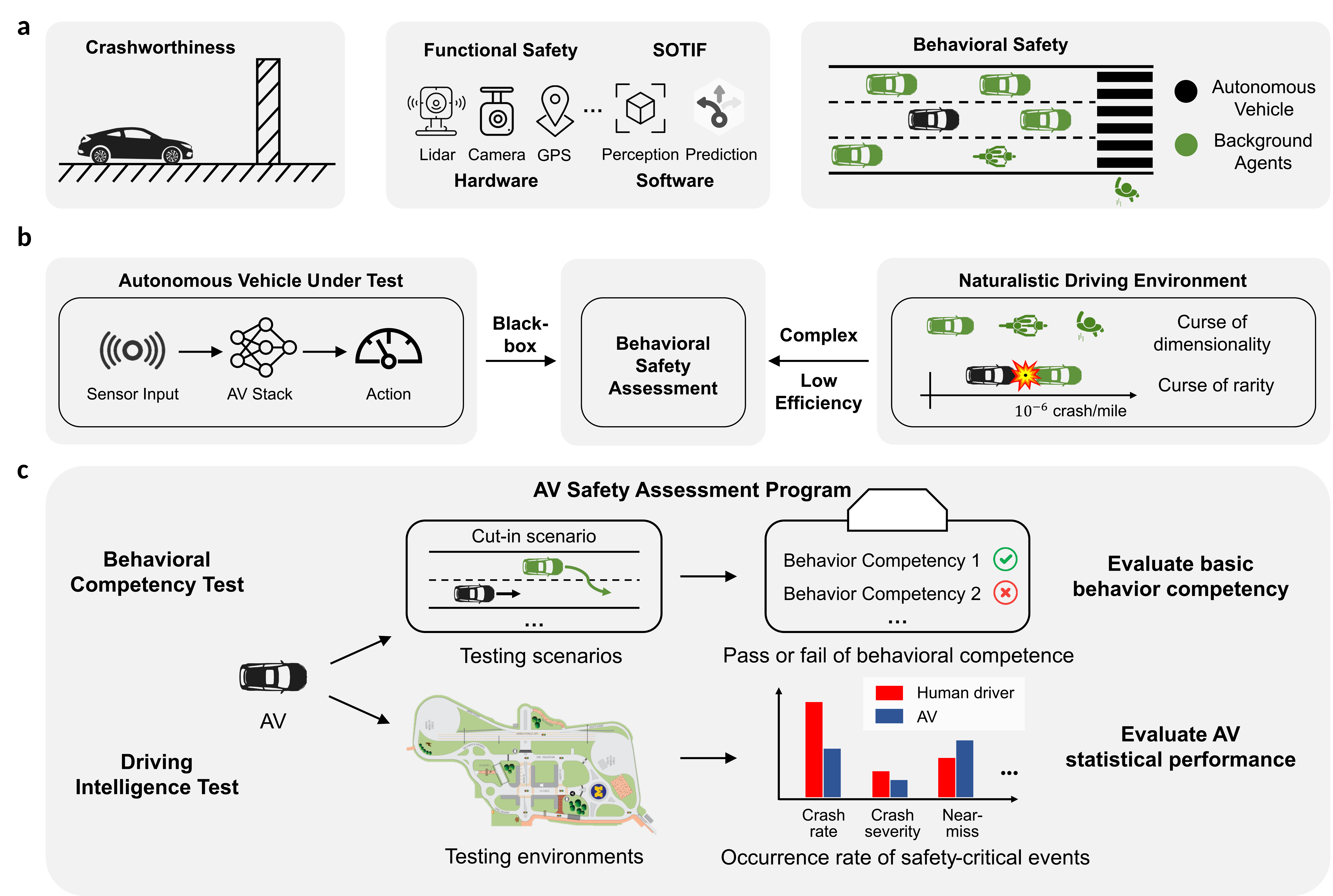}
  \caption{Overview of the behavioral safety assessment framework. (a) Distinction between behavioral safety, crashworthiness, and functional safety. (b) Key challenges in third-party AV behavioral safety evaluation. (c) The proposed framework consisting of the Behavioral Competency Test (BCT) and the Driving Intelligence Test (DIT).}
  \label{fig:framework}
\end{figure*}

The above discussions underscore the necessity of shifting the AV evaluation paradigm from functional safety to behavioral safety, which evaluates whether a functionally sufficient AV can make safe decisions in reacting to and interacting with the traffic environment, without creating or exacerbating safety-critical situations.
Consequently, behavioral safety is fundamentally distinct from crashworthiness and functional safety, as shown in Fig.~\ref{fig:framework}a.
The CertiCAV Assurance Paper~\cite{cpc_certicav} by Connected Places Catapult of the United Kingdom also highlights the importance of behavioral safety, reinforcing our argument that functional safety alone is insufficient for AV evaluation.
However, it falls short of providing concrete definitions and practical methodologies for implementation, leaving a critical gap in the current AV testing approaches.
In recent years, while some AV companies have released information about their internal testing procedures~\cite{favaro2023building,cruise2023safety,aurora_safety} related to behavioral safety, these disclosures remain limited, and the specifics of their testing protocols are often inaccessible.
Therefore, developing a systematic, transparent, and robust third-party testing framework for the behavioral safety of AVs is an urgent priority.

Conducting third-party evaluations of AV behavioral safety presents significant challenges, as illustrated in Fig.~\ref{fig:framework}b. First, modern AVs, particularly end-to-end models, lack interpretability and operate as black-box systems, making failures unpredictable.
This means failures may occur not only in rare edge cases but also during routine driving.
Such unpredictability erodes public trust, especially when crashes happen under normal conditions.
A recent study shows that over 90\% of AV crashes occur in situations with no unusual road conditions or dangerous pre-accident behaviors~\cite{abdelaty2024matched}, contradicting the common belief that AV testing should focus primarily on rare edge cases.
This highlights the need for a holistic evaluation of basic behavioral competency rather than just extreme-case testing.
Second, AV testing is inherently difficult due to the complexity of real-world traffic and the rarity of safety-critical events. The naturalistic driving environment (NDE) involves intricate spatial-temporal interactions among diverse road users, across varying road geometries, and under dynamic traffic conditions~\cite{yan2023learning}.
Accurately replicating these factors in simulations requires high-fidelity modeling of both physical environments and human behaviors, making AV assessment highly complex.
Compounding this challenge is the curse of rarity~\cite{liu2024curse}: safety-critical events occur infrequently, requiring billions of miles to collect sufficient data for meaningful evaluation~\cite{feng2023dense}.
This renders brute-force mileage-based testing impractical, even in simulations, where the key challenge is generating safety-critical events both realistically and efficiently.

To address these challenges, we propose a dedicated third-party testing initiative for systematically evaluating AV behavioral safety.
The framework consists of two key components: the Behavioral Competency Test (BCT) and the Driving Intelligence Test (DIT), as illustrated in Fig.~\ref{fig:framework}c.
The BCT evaluates AV performance in predefined scenarios of varying risk levels, akin to human driver's license tests, to verify basic behavioral competency in both normal and safety-critical situations.
The DIT, in contrast, goes beyond known scenarios by constructing a Naturalistic and Adversarial Driving Environment (NADE) that leverages importance sampling to efficiently uncover unknown failure modes and statistically measure safety-critical event rates without bias. Together, the BCT ensures baseline competency in expected driving situations, while the DIT reveals long-tail risks that may only emerge at scale.
Besides, both of them operate in a black-box manner, requiring no knowledge of the AV's internal mechanisms, thus ensuring fairness and objectivity.

This study is presented in two parts.
This paper (Part I) establishes the overall framework and methodologies.
Section~\ref{sec: related work} reviews related work and identifies existing gaps.
Section~\ref{sec: methodology} presents the proposed methodology: Section~\ref{subsec: framework} defines behavioral safety and introduces the assessment framework, Section~\ref{subsec: formulation} formulates the problem, and Sections~\ref{sec: bct} and~\ref{sec: dit} detail the BCT and DIT, respectively.
Section~\ref{sec: evaluation} introduces the performance evaluation method.
Part II of this study~\cite{Liu2026partII} validates the proposed framework through both simulation and field test results.

\section{Related Work}
\label{sec: related work}

In this section, a brief overview of related work is provided from two areas: the well-established testing programs and standards, and existing work on behavioral safety evaluation.

There exist several well-established testing programs for conventional human-driven vehicles~\cite{nhtsa_ncap,euroncap,iihs_ratings,cncap}. For example, the New Car Assessment Program (NCAP)~\cite{nhtsa_ncap} is a safety rating program that primarily assesses vehicle crashworthiness and crash avoidance technologies such as Autonomous Emergency Braking (AEB). Similar programs have been adopted worldwide, including Euro NCAP~\cite{euroncap} in Europe and C-NCAP~\cite{cncap} in China, while the Insurance Institute for Highway Safety (IIHS)~\cite{iihs_ratings} provides complementary vehicle safety evaluations in the United States. These programs all follow a similar paradigm focused on whether a specific passive or active safety feature performs correctly in predefined test scenarios rather than on evaluating a vehicle's overall driving behavior, which is insufficient for Level 4--5 AVs.
For high-level AVs, notable standards include ISO 26262~\cite{iso26262} for FuSa, ISO 21448~\cite{iso21448} for SOTIF, and ISO/PAS 8800~\cite{iso8800} for Road Vehicles Safety and Artificial Intelligence, all of which address risks arising from hardware and software malfunctions or insufficiencies. In addition, ISO 34502~\cite{iso34502} introduces a scenario-based testing framework for SOTIF, while UL 4600~\cite{ul4600} provides methods for assessing AV's overall safety case. However, these testing standards adopt a vehicle-centric perspective, assessing AV safety based on vehicle system function and capability, such as sensor failures (hardware) and perception errors (software).
These vehicle-centric approaches are insufficient for evaluating AV safety in large-scale real-world deployments, as we discussed in Section~\ref{sec: introduction}.

Most of the existing AV safety assessment methods, including the scenario-based methods~\cite{feng2020testingI,feng2020testingII,zhong2022neural,zhou2023specification} and environment-based methods~\cite{feng2021intelligent,feng2023dense,yan2023learning}, are not designed to evaluate behavioral safety.
Several researchers have also recognized the absence of rigorous AV behavioral safety evaluation.
In the CertiCAV Assurance Paper~\cite{cpc_certicav} by Connected Places Catapult of the United Kingdom, behavioral safety is defined as ``the field of safety concerned with addressing the safety and sufficiency of the intended functionality of driving features.''
Its objectives include traffic rule compliance, active/tactical safety controls, and fallback behavior.
This definition, however, focuses on whether the intended functionality is safe and sufficient, and therefore remains oriented toward the specification and development process.
It falls short of providing practical testing methodologies for evaluating actual AV behavioral performance in traffic, leaving a critical gap in existing AV testing approaches.
In recent years, some AV companies have disclosed information about their internal testing procedures related to behavioral safety~\cite{favaro2023building,cruise2023safety,aurora_safety}.
Waymo published their approach for determining the absence of unreasonable risk, structured around architectural hazards, behavioral hazards, and in-service operational hazards~\cite{favaro2023building}.
Although behavioral hazards bear some resemblance to behavioral safety, they are treated as one component of the AV development life cycle, decomposing the AV system and evaluating each subsystem individually.
This approach is inconsistent with our goal of treating the AV as a black box from a third-party evaluation perspective.
Cruise reported coverage of SOTIF, FuSa, physical safety/crashworthiness, passenger safety, and operational safety~\cite{cruise2023safety}, which is distinct from behavioral safety as defined in this work.
Similarly, Aurora grounded their behavioral safety processes in SOTIF and FuSa~\cite{aurora_safety}.
Overall, these industry disclosures remain limited in scope, and the specifics of their testing protocols are largely inaccessible to the public.

\section{Behavioral Safety Assessment Method}
\label{sec: methodology}

\begin{figure*}[t]
  \centering
  \includegraphics[width=\textwidth]{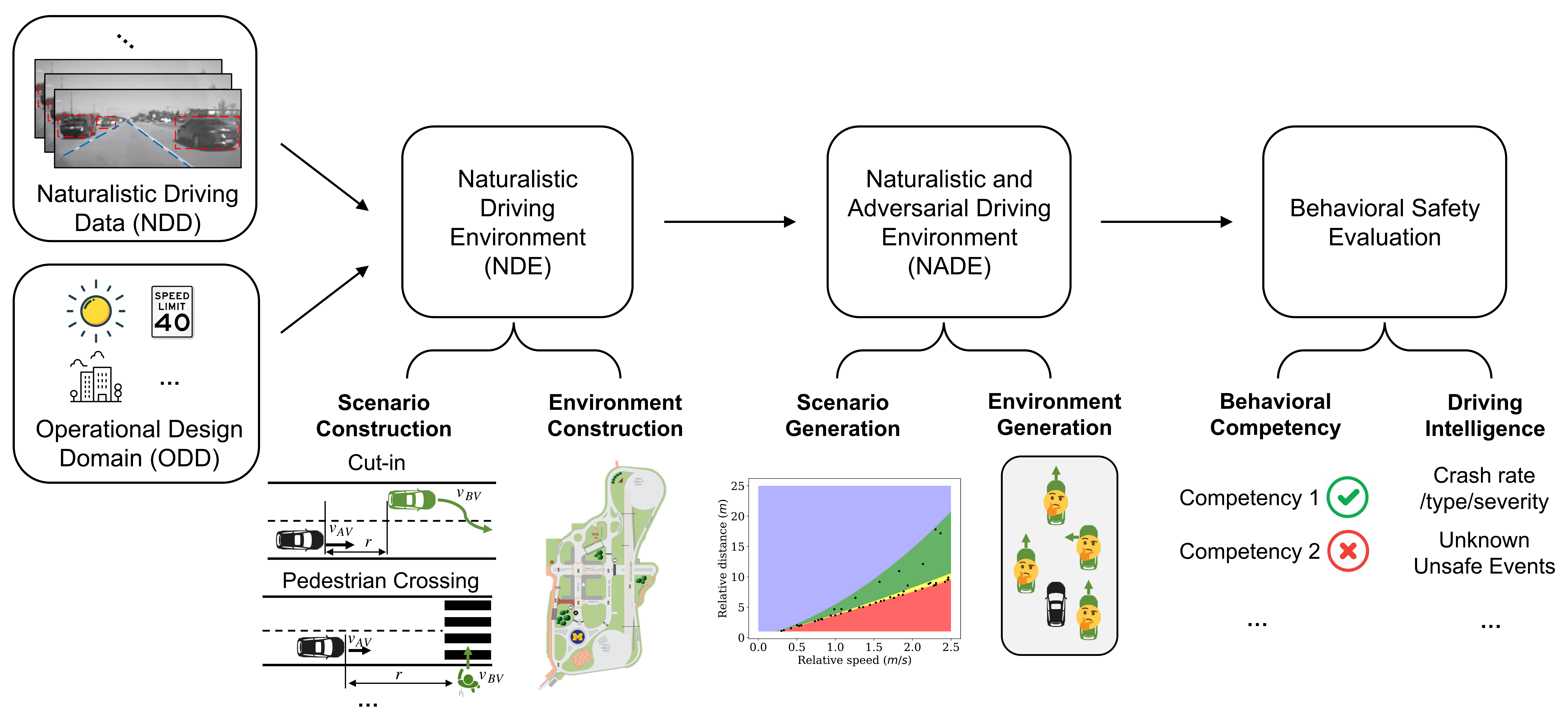}
  \caption{Overall pipeline of developing the Behavioral Competency Test and the Driving Intelligence Test}
  \label{fig: pipeline}
\end{figure*}

\subsection{Behavioral safety assessment framework}
\label{subsec: framework}

As discussed in Section~\ref{sec: introduction}, behavioral safety aims to evaluate an AV's ability to mitigate or avoid hazards and not create new safety hazards without being solely based on predetermined scenarios, even though the AV is fully functional.
Herein, we propose the definition of behavioral safety:
\begin{definition}
    Behavioral safety refers to the absence of unreasonable risk due to hazards resulting from an AV's unsafe behavioral responses to and interaction with the traffic environment.
\end{definition}
Specifically, it encompasses two key aspects of unsafe AV decision-making: (1) reactive behavior, which involves failures in avoiding or mitigating hazards when presented with safety-critical situations, and (2) interactive behavior, which concerns both an AV's potential to unintentionally expose itself to safety-critical situations and its inability to minimize the likelihood of secondary incidents.
To illustrate behavioral safety, consider a scenario where a human-driven background vehicle (BV) aggressively cuts into the AV's lane. Even if the AV system is fully functional and detects the maneuver accurately, behavioral safety evaluation assesses the AV's capability to make appropriate decisions, such as timely braking or performing an evasive lane change, without causing another safety-critical situation.
Behavioral safety also includes proactive risk mitigation measures, such as avoiding prolonged driving in another vehicle's blind spot to reduce potential cut-in incidents. Furthermore, behavioral safety explicitly evaluates the repercussions of an AV's interactive behavioral decisions in preventing secondary incidents.
For instance, when an AV executes a hard braking maneuver, the behavioral safety assessment must consider whether this decision could inadvertently cause a rear-end collision with a following BV.
Ultimately, behavioral safety evaluates whether an AV can intelligently react to and interact with both static road infrastructure and dynamic road users when the system is functionally sufficient and free of errors.

The framework consists of two key components: the Behavioral Competency Test (BCT) and the Driving Intelligence Test (DIT), as illustrated in Fig.~\ref{fig:framework}c.
The BCT focuses on evaluating AV performance in ``known'' scenarios, assessing basic behavioral competency based on human prior knowledge.
Inspired by human driver's license tests, it examines fundamental driving skills across both normal and safety-critical situations, such as car-following and cut-ins.
Each driving skill is evaluated across multiple test cases with varying risk, where adversarial background agents are introduced in high-risk cases to provoke challenging situations, enabling assessment of the AV's reactive behavior in avoiding and mitigating hazards.
This approach enables a broad evaluation beyond rare edge cases, providing a baseline assessment of AV behavioral safety in situations where human drivers are expected to perform reliably.
It aligns conceptually with existing scenario-based testing practices commonly adopted by testing organizations~\cite{tuvsud_scenario} and by AV companies~\cite{tesla_fsd_testing}.

In contrast, the DIT uncovers ``unknown unsafe'' events through statistical evaluation, providing a comprehensive assessment of AV behavioral safety within its Operational Design Domain (ODD) by considering both reactive and interactive behaviors. Instead of relying on predefined scenarios, the DIT constructs a high-fidelity NDE to create a realistic background traffic environment for AV assessment. To overcome the curse of rarity, we introduce the Naturalistic and Adversarial Driving Environment (NADE), which builds upon NDE to intelligently increase the exposure frequency of safety-critical events using importance sampling theory. This enables the DIT to measure the occurrence rate of safety-critical events (e.g., crashes) involving both AVs and BVs, accelerating testing without introducing bias, thus significantly improving efficiency while maintaining statistical rigor. Beyond statistical performance measurement, a key advantage of the DIT is its ability to reveal previously unknown failure modes, which is particularly crucial before large-scale AV deployments. Together, BCT and DIT form a complementary framework, addressing both known and unknown sources of AV unpredictability. The BCT ensures AVs possess essential behavioral competencies expected of human drivers, while the DIT identifies long-tail risks that may only emerge at scale. A key strength of our framework is its testing-oriented design, which treats the AV as a black box, ensuring fairness and objectivity by not requiring prior knowledge of its internal mechanisms.

The design process for both the BCT and DIT follows a unified framework, as illustrated in Fig.~\ref{fig: pipeline}.
The first step involves creating an NDE based on Naturalistic Driving Data (NDD) and the specific ODD.
The NDE forms the foundation for behavioral assessment, simulating real-world traffic conditions.
However, directly generating and sampling testing environments from the NDE can be inefficient~\cite{feng2023dense,kalra2016driving}, as safety-critical events are rare in real-world traffic~\cite{liu2024curse}, rendering many environments irrelevant for safety assessments.
To address this, we introduce the NADE, which builds upon the NDE to create and sample testing scenarios and environments more effectively.
The key idea of the NADE is to intensify the occurring frequencies of safety-critical events while maintaining comprehensive coverage for the BCT and ensuring unbiased results in the DIT.
It is important to note that the difference between a ``scenario'' and an ``environment'' largely pertains to scale and complexity: environments encompass broader spatiotemporal coverage, with a larger number of background road users.
Various scenarios, featuring different types of interactions, can emerge spontaneously and dynamically within the environment.
In this sense, a scenario can be viewed as a simplified version of an environment, which is why we refer to the term ``environment'' in the NDE and NADE throughout this paper.

\subsection{Formulation of AV behavioral safety assessment problem}
\label{subsec: formulation}

This section describes the formulation of AV behavioral safety assessment problem.
Denote the variable of the driving environment as $\boldsymbol{x} = [\boldsymbol{s}(0),\boldsymbol{u}(0),\boldsymbol{u}(1),\ldots,\boldsymbol{u}(k)]$, in which $\boldsymbol{s}(k)$ denotes the state vector (position, speed, heading, etc.) of the AV and BVs at the $k$-th timestep, and $\boldsymbol{u}(k)$ denotes the action vector of BVs at this step. The core objective of behavioral safety evaluation is to determine whether an AV system can operate safely within this dynamic driving environment. Let $\pi$ denote the AV model and its safety performance $\sigma_{\pi}$ can be measured by
\begin{equation}
\sigma_{\pi} = \mathcal{F}(\boldsymbol{x}, \pi), \quad \boldsymbol{x} \sim P(\boldsymbol{x}),
\end{equation}
where $\mathcal{F}(\cdot)$ denotes testing methods and procedures that derive AV safety performance from the driving environment $\boldsymbol{x}$, and $P(\boldsymbol{x})$ denotes the distribution of the NDE.

Behavioral safety is dependent on two critical aspects: (1) the AV's behavioral capability when confronted with potentially hazardous situations, and (2) the AV's ability to avoid entering such dangerous situations altogether.
The first aspect addresses the AV's performance in safety-critical events triggered by itself or other agents, such as a sudden cut-in by an adjacent vehicle.
The second aspect evaluates the AV's overall competence of navigating the environment.
It involves not only the capability of reacting to dangerous situations but more importantly its capabilities of reducing the chances of posing itself in dangerous situations, for example, not staying in the blind spot of other vehicles to reduce the chances of being cut in.

To comprehensively evaluate these two aspects, we propose a dual testing framework consisting of the BCT and the DIT. In the BCT, the constraints of the traffic environment are predefined to establish a specific testing scenario.
The testing result for each testing episode is binary, indicating whether a safety-critical event occurs or not:
\begin{equation}
\sigma_{\pi}(\boldsymbol{x}) = \mathbb{I}(A \mid \boldsymbol{x}, \pi), \quad \boldsymbol{x} \sim P(\boldsymbol{x}),
\end{equation}
where $A$ denote the safety-critical event, for example, crash event is used in this study, $\mathbb{I}(\cdot)$  is the indicator function which equals to one if a crash happens during the testing case $\boldsymbol{x}$ and equals zero otherwise, and $P(\boldsymbol{x})$ here denotes the distribution of testing cases in the scenario set. Based on the pass and fail results of AV under diverse testing cases, we can evaluate its behavioral competence of handling dynamic driving tasks within the ODD.

The Driving Intelligence Test is designed to assess statistical safety performance of AV, which can be measured by
\begin{equation}
\sigma_{\pi} = P(A) = \mathbb{E}\left(P(A|\boldsymbol{x})\right), \quad \boldsymbol{x} \sim P(\boldsymbol{x}),
\end{equation}

where $P(A)$ denotes the crash rate of AV. The testing process is to deploy the AV in the simulated NDE, observe its safety performance during the dynamic interaction with background agents, and obtain the occurrence rate of safety-critical events by accumulating a large amount of testing episodes. This evaluated crash rate demonstrates the overall safety performance of AVs and provides statistical evidence of AV behavioral safety during real-world traffic deployments.

As safety-critical events are rare events in the real-world traffic environment, most driving scenarios and environments $\boldsymbol{x}$ in both BCT and DIT would be not challenging for AV. As a result, it will take large number of tests to obtain the evaluation results~\cite{feng2023dense,kalra2016driving}, which is time-consuming and even impractical for real-world tests. To improve testing efficiency, the key is to generate a new testing environment distribution $q(\boldsymbol{x})$, we termed it Naturalistic and Adversarial Driving Environment (NADE), in which safety-critical events are intelligently identified and generated to increase the exposure of AV to these high-valued events. Therefore, for both BCT and DIT, the testing environment will be sampled from the NADE rather than NDE, aiming to evaluate the AV behavioral safety with high accuracy and efficiency.

In summary, both the BCT and the DIT follow the same overall framework, first constructing the naturalistic environment distributions $P(\boldsymbol{x})$  from real-world datasets, then generating an intelligent environment distribution $q(\boldsymbol{x})$ to improve testing efficiency. The BCT is a simplified case of the DIT, with a smaller number of background agents and shorter time intervals. The detailed processes of constructing both tests are discussed in the following sections.

\begin{figure*}[t]
  \centering
  \includegraphics[width=\textwidth]{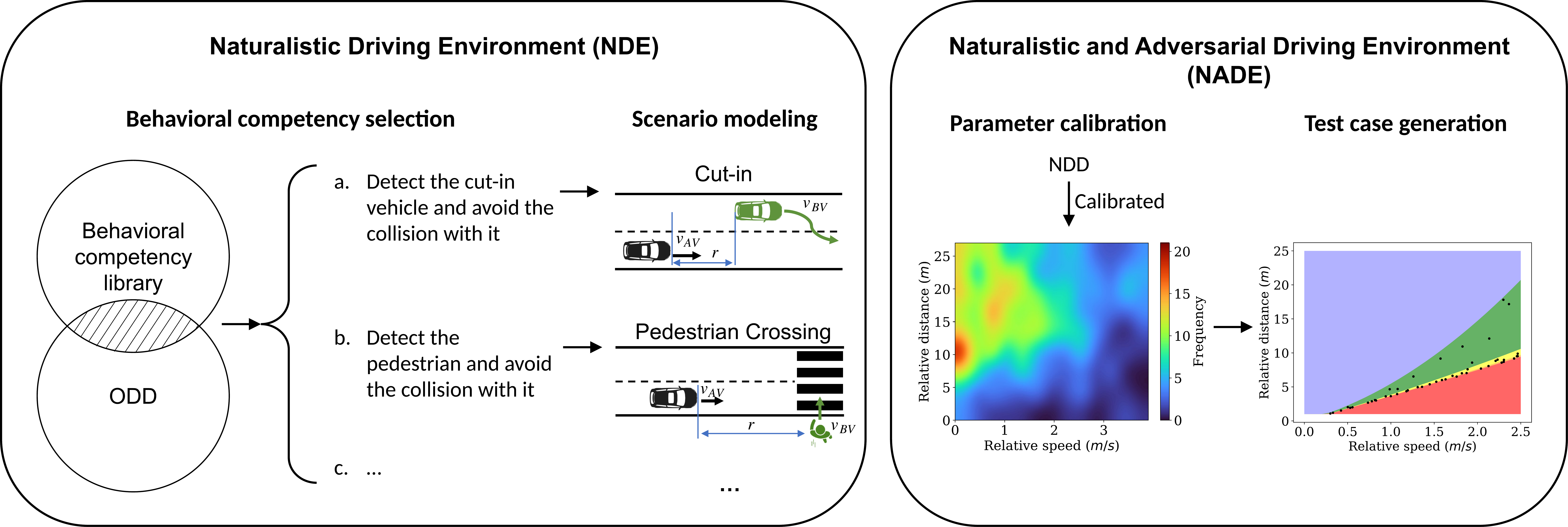}
  \caption{The components of developing the BCT}
  \label{fig: bct pipeline}
\end{figure*}

\section{Behavioral Competency Test}
\label{sec: bct}

As discussed in Section~\ref{subsec: formulation}, the target of BCT is to evaluate whether the AV possesses the basic behavioral competency.
Here, we will use the definition of behavioral competency proposed in Ref.~\cite{nhtsa2017ads}: behavioral competency refers to the ability of an ADS to operate in the traffic conditions that it will regularly encounter, including keeping the vehicle in a lane, obeying traffic laws, following reasonable road etiquette, and responding to other vehicles or hazards.
To achieve this goal, the BCT is developed in two steps: NDE construction (scenario construction) and NADE generation (scenario generation), illustrated in Fig.~\ref{fig: bct pipeline}.

\subsection{NDE construction for BCT}
\label{subsec: bct nde}

To construct the NDE for the BCT, the first step is to identify the behavioral competencies to be evaluated.
The selected behavioral competencies are determined by the intersection of the basic behavioral competency library and the ODD of the AV deployment area, as illustrated by the hatched region in the left box of Fig.~\ref{fig: bct pipeline}.
Generally, each behavioral competency corresponds to one test scenario.
Two examples are shown in Fig.~\ref{fig: bct pipeline}, where the first behavioral competency, ``Detect the cut-in vehicle and avoid the collision with it'', corresponds to the cut-in scenario, and the second behavioral competency, ``Detect the pedestrian and avoid the collision with it'', corresponds to the pedestrian crossing scenario.

The next step is scenario modeling, in which we parameterize each scenario.
Generally, a scenario can be described by a constrained dynamical system:
\begin{equation}
\left\{
\begin{aligned}
  & \dot{\boldsymbol{s}}(t) = f(\boldsymbol{s}(t), \boldsymbol{u}(t)), \\
  & (\boldsymbol{s}(t), \boldsymbol{u}(t)) \in \mathcal{C}(t)
\end{aligned}
\right.
\end{equation}
where $\mathcal{C}(t) = \{ (\boldsymbol{s}(t), \boldsymbol{u}(t)) | h((\boldsymbol{s}(t), \boldsymbol{u}(t)), t) = 0 \}$ represents the constraints, $h(\cdot)$ is the constraint function.
Different constraints correspond to different test cases.
We take the cut-in scenario as an example to show how the model works.
The state vector is defined as follows:
\begin{equation}
    \boldsymbol{s} = [\Delta d, \Delta v]^T
\end{equation}
where $\Delta d$ is the relative longitudinal distance of AV and BV, $\Delta v$ is the relative longitudinal speed of AV and BV.
The control vector is defined as follows:
\begin{equation}
    \boldsymbol{u} = [a, \omega]^T
\end{equation}
where $a$ is BV's acceleration, and $\omega$ is BV's yaw rate.
For easy execution in physical tests, we assume that the BV speed is constant and will not react to the behavior of AV after the BV starts to have lateral movement in this study.
Therefore, the scenario is uniquely determined given the relative distance $\Delta d(0)$ and relative speed $\Delta v(0)$ at the moment when BV starts to cut-in.
The constraint function is defined as:
\begin{equation}
    h(\boldsymbol{s}(t), \boldsymbol{u}(t), t) = 
    \begin{bmatrix}
        \Delta d(0) - \Delta d_0 \\
        \Delta v(0) - \Delta v_0 \\
        a(t)
    \end{bmatrix}
    = \boldsymbol{0}
\end{equation}
where $a(t) = 0$ enforces the constant-speed assumption of the BV.
For different $\Delta d_0$ and $\Delta v_0$, we can obtain different test cases.
The constraints naturally differ across scenarios.
For example, they are the initial relative distance between AV and vulnerable road user (VRU), and VRU speed in the VRU Crossing the Street Without Crosswalk scenario, while in the traffic light scenario, it only consists of the time-headway (THW) from the AV to the stop line.
The detailed constraint settings can be found in Part II of this study~\cite{Liu2026partII}.
Note that the inputs of the constraints include but are not limited to the physical variables, surrogate model, or distributions.
And the control command of BV can also be calculated by using optimal control theory, neural network, etc.

\subsection{NADE generation for BCT}
\label{subsec: bct nade}

To accelerate the testing process, we first calibrate each scenario by fitting the state distribution to human driving data, thereby constructing a naturalistic distribution $P(\boldsymbol{x})$.
This calibration step is essential to ensure that the scenarios are representative of situations the AV will encounter during deployment, and to exclude infeasible parameter settings in which collisions are unavoidable or the required vehicle behavior exceeds physical limits.
Naturalistic driving data (NDD) is leveraged to extract vehicle trajectories for calibrating each scenario.
Since different test cases are obtained by specifying different constraint parameters, the calibration focuses on estimating the distribution of those constraint-relevant parameters.
The left image in the right box of Fig.~\ref{fig: pipeline} shows an example of the calibration results of the cut-in scenario.
The x axis and y axis are the initial relative speed and initial relative distance, respectively, which are used to define the constraints.

The final step is test case generation, which determines concrete test cases for each scenario. Since the goal of BCT is to evaluate whether an AV possesses a certain behavioral competency, meaning that it will not collide in any possible situation within the scenario, we do not randomly sample from $\mathrm{P}(\boldsymbol{x})$ based solely on occurrence rates, nor do we perform importance sampling to estimate a crash rate. Instead, we seek to certify that the AV avoids collision across the full constraint space with high confidence.
Since testing all possible situations is infeasible, we balance coverage and risk across the testing space, so that when an AV passes all sampled test cases, we can conclude that it avoids collision in a certified proportion of situations, that is, it possesses the behavioral competency at a quantifiable confidence level.

To achieve this, we assume that if the AV passes a test case with constraint $\boldsymbol{c}_0$, it
will also pass any test case whose constraint $c$ lies within a neighborhood of $\boldsymbol{c}_0$
with high probability.
The harder the test case, the smaller this neighborhood.
Formally, the risk of a test case under constraint $\boldsymbol{c}$ is denoted $\delta(\boldsymbol{c})$,
and the neighborhood radius is defined as its inverse:
\begin{equation}
    \gamma(\boldsymbol{c}) = \frac{1}{\delta(\boldsymbol{c})}
\end{equation}

Under this assumption, with constraint $\boldsymbol{c}_0 \in \mathcal{C}$, the conditional probability
of failure at a nearby test case with constraint $\boldsymbol{c} \in B(\boldsymbol{c}_0, \gamma(\boldsymbol{c}_0))$ is denoted
$p(A \mid \boldsymbol{c}, \boldsymbol{c}_0)$, where $B(\boldsymbol{c}_0, \gamma(\boldsymbol{c}_0))$ is the neighborhood of $\boldsymbol{c}_0$ with radius
$\gamma(\boldsymbol{c}_0)$.
To reflect that lower-risk regions warrant broader coverage, we define a
radius-weighted local sampling density for each test case $\boldsymbol{c}_i$, normalized over its neighborhood:
\begin{equation}
    q_i(\boldsymbol{c}) = \frac{\gamma(\boldsymbol{c})}{\displaystyle\int_{B(\boldsymbol{c}_i,\,\gamma(\boldsymbol{c}_i))} \gamma(\boldsymbol{c}')\,\mathrm{d}\boldsymbol{c}'}
\end{equation}
Given a single passed test case $\boldsymbol{c}_i$, the probability of no hidden failure within its neighborhood is:
\begin{equation}
    P_i = 1 - \int_{B(\boldsymbol{c}_i,\,\gamma(\boldsymbol{c}_i))} p(A \mid \boldsymbol{c}, \boldsymbol{c}_i)\, q_i(\boldsymbol{c})\,\mathrm{d}\boldsymbol{c}
\end{equation}

For $n$ passed test cases, combining over the full covered region
$\mathcal{C} = \bigcup_{i=1}^{n} B(\boldsymbol{c}_i, \gamma(\boldsymbol{c}_i))$, the test environment distribution is normalized over $\mathcal{C}$:
\begin{equation}
    q(\boldsymbol{x}) = q(\boldsymbol{c}) = \frac{\gamma(\boldsymbol{c})}{\displaystyle\int_{\mathcal{C}} \gamma(\boldsymbol{c}')\,\mathrm{d}\boldsymbol{c}'}
\end{equation}
The probability of failure at a test case with constraint $\boldsymbol{c}$ decreases as more neighboring test cases are passed and as the test case becomes easier:
\begin{equation}
    p(A \mid \boldsymbol{c}, \{\boldsymbol{c}_i\}_{i=1}^{n}) \propto \delta (\boldsymbol{c}) \prod_{i:\, \boldsymbol{c} \in B(\boldsymbol{c}_i,\, \gamma(\boldsymbol{c}_i))} p(A \mid \boldsymbol{c}, \boldsymbol{c}_i)
\end{equation}
The overall confidence is then:
\begin{equation}
    \mathrm{Conf}(n) = 1 - \int_{\mathcal{C}}
    p(A \mid \boldsymbol{c}, \{\boldsymbol{c}_i\}_{i=1}^{n})q(\boldsymbol{c})\,\mathrm{d}\boldsymbol{c}
\end{equation}

If the AV passes all $n$ test cases, it is certified to possess the behavioral competency at confidence level $\mathrm{Conf}(n)$. The minimum number of test cases required to reach a target confidence level $P^*$ is therefore:
\begin{equation}
    n^{*} = \min\left\{ n \in \mathbb{N} : \mathrm{Conf}(n) > P^{*} \right\}
    \label{equ: sampling number}
\end{equation}

As an example, in the cut-in scenario, we first determine if a case is hard or not by calculating the required minimum deceleration of the AV to avoid the collision.
The larger the value is, the riskier the case.
Based on the value and NDD, we divide cases into five different risk levels, that is, trivial, low, medium, high, and infeasible risk levels.
And the risks are set to different constants at different risk levels.
For the trivial level, where test cases are considered very safe and require minimal or no action from the vehicle, the risk is set to 0, which means we won't sample test case from this level.
And for the infeasible level, where collisions are unavoidable, the risk is set to infinity, which means we won't sample test cases from this level, either, as it is impossible to sample test cases with infinite sampling density.
The probability of no hidden failure within a passed test case is set to 1.
As a result, the state space segmented with risk levels constitutes the new sampling distribution $q(\boldsymbol{x})$, as shown in the right figure in the NADE box in Fig.~\ref{fig: bct pipeline}, covering the state space and diverse risk levels.
In this figure, the trivial, low, medium, high, and infeasible risk levels are represented by the blue, green, yellow, orange, and red areas, and the black points are sampled test cases.

\section{Driving Intelligence Test}
\label{sec: dit}

\begin{figure*}[t]
  \centering
  \includegraphics[width=\textwidth]{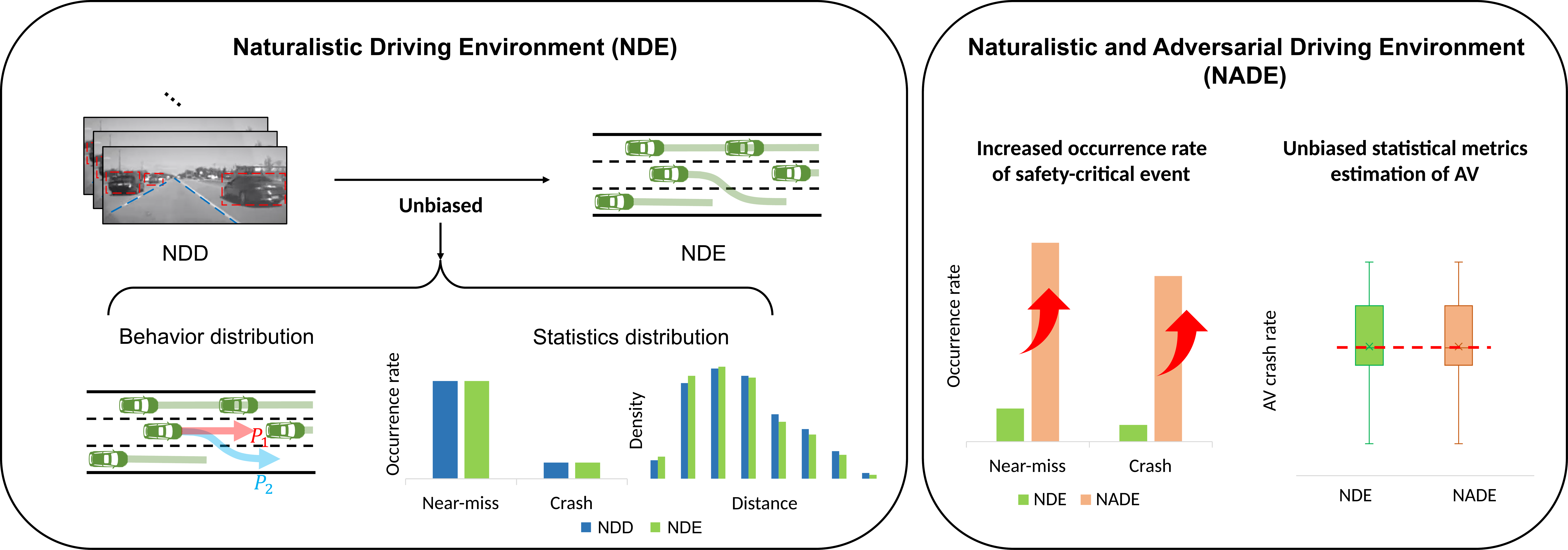}
  \caption{The components of developing the DIT}
  \label{fig: dit pipeline}
\end{figure*}

The DIT aims to evaluate the crash rate of AV.
It also consists of NDE construction (environment construction) and NADE generation (environment generation), illustrated in Fig.~\ref{fig: dit pipeline}.

\subsection{NDE construction for DIT}
\label{subsec: dit nde}

To obtain an unbiased estimate of AV crash rates, it is essential to construct a high-fidelity driving environment. Specifically, both the behavioral distribution $P(\boldsymbol{u}|\boldsymbol{s})$ and safety-critical event distribution $P(A)$ in the NDE should match those in NDD, as illustrated in the left box of Fig.~\ref{fig: dit pipeline}.
To model human driving behavior and learn its underlying distribution, a wide range of approaches have been proposed, including model-based car-following~\cite{treiber2000-IDM} and lane-changing models~\cite{kesting2007-MOBIL}, Markovian-based methods~\cite{chen2010-CF-model, yan2025distributionally-nde}, as well as recent learning-based and generative AI approaches~\cite{suo2021-trafficsim, yan2023learning, nips24-smart, xiao2026long}. In this study, we adopt the method developed in~\cite{feng2021intelligent, yan2025distributionally-nde}.

The next step is to validate $P(A)$ of the NDE.
By using the Monte Carlo method, $P(A)$ can be derived as:
\begin{equation}
P(A) = \mathbb{E}_{\boldsymbol{x} \sim P(\boldsymbol{x})}\left(P(A|\boldsymbol{x})\right) \approx \frac{1}{n}\sum_{i=1}^{n} P(A|\boldsymbol{x}_i), \quad \boldsymbol{x}_i \sim P(\boldsymbol{x}),
\end{equation}
where $n$ is the number of testing episodes, and $i$ denotes the $i$-th test.
The actions of BVs $\boldsymbol{u}(k)$, conditioned on the current state $\boldsymbol{s}(k)$, follow a realistic distribution $P(\boldsymbol{u}(k)|\boldsymbol{s}(k))$.
With the Markovian assumptions of BV behavior, the naturalistic probability of each testing episode can be calculated as
\begin{equation}
P(\boldsymbol{x}) = P(\boldsymbol{s}(0)) \cdot \prod_{k=0}^{T} P(\boldsymbol{u}(k)|\boldsymbol{s}(k)).
\end{equation}

Although the NDE reproduces realistic driving behavior, the occurrence rate of safety-critical events may not align with real-world statistics due to their inherent rarity in naturalistic driving data~\cite{liu2024curse}.
To address this gap, the generation of safety-critical events within the DIT framework is grounded in human error modeling, drawing on empirical studies of the fundamental causes of vehicular crashes.
Following the taxonomy proposed by Ref.~\cite{singh_critical_2015}, errors are categorized into recognition (41\% $\pm$ 2.2\%), decision (33\% $\pm$ 3.7\%), performance (11\% $\pm$ 2.7\%), and sleep-related factors (7\% $\pm$ 1.0\%).
In our implementation, each vehicle is assigned a probability $p$ of neglecting one of its six nearest neighbors for a duration $t$ at each timestep. This mechanism allows for the emergence of diverse safety-critical events such as dangerous cut-ins, red-light violations, unprotected left turns, and roundabout conflicts.

To ensure that the simulated distribution of crash types aligns with real-world statistics, we introduce a calibration mechanism that adjusts the occurrence probabilities of human error sources. Each human error type $i$ is associated with a base probability $p_i$, and each crash type $k$ is affected by multiple such errors. Based on empirical crash causation studies, we define a matrix $A$ where each element $A_{k,i}$ represents the contribution of error $i$ to the observed frequency of crash type $k$.

During the calibration, we apply a multiplicative influence factor $I_i$ to each $p_i$, yielding an adjusted probability $p_i' = I_i \cdot p_i$. We adopt a simplifying but necessary assumption: the crash rate induced by a particular human error is linearly proportional to its occurrence probability. Under this assumption, the simulated frequency of crash type $k$ is given by:
\begin{equation}
    \hat{R}_k = \sum_i A_{k,i} \cdot I_i.
\end{equation}

Given the empirical target distribution $R_k$, we seek the set of influence factors $I_i$ that minimizes the overall deviation between simulated and real-world crash frequencies. Introducing slack variables $t_k$ to represent the absolute deviation for each crash type, we formulate the following optimization problem:
\begin{equation}
    \begin{aligned}
    \min_{I_i, t_k} \quad & \sum_k t_k \\
    \text{subject to} \quad 
    & -t_k \leq \sum_i A_{k,i} I_i - R_k \leq t_k, \quad && \forall k, \\
    & 0.5 \leq I_i \leq 2, \quad && \forall i.
    \end{aligned}
\end{equation}
where, $I_i$ directs how much the probability of each human error is scaled up or down to improve alignment with empirical crash data. The bounds on $I_i$ constrain the adjustment range, preventing unstable or unrealistic behavior during optimization.

Since the assumption of linear proportionality between crash rates and $p_i$ holds only locally, we adopt an iterative optimization strategy. In each round, $I_i$ is adjusted within the bounded range $[0.5, 2]$, followed by a new run and recalibration. This iterative refinement ensures convergence toward a stable accident distribution that closely matches real-world statistics, while maintaining fidelity in behavior-level dynamics.

\subsection{NADE generation for DIT}
\label{subsec: dit nade}

Testing efficiency in the NDE is inherently limited by the rarity of safety-critical events.
To improve efficiency, the occurrence rate of safety-critical events must be substantially increased, while ensuring that the resulting estimate of the AV's crash rate remains unbiased, as illustrated in the right box of Fig.~\ref{fig: dit pipeline}.
To achieve this, we developed the NADE based on NDE, forming an intelligent testing environment $q(\boldsymbol{x})$ to evaluate AV performance based on importance sampling theory~\cite{owen2013monte}
\begin{equation}
P(A) \approx \frac{1}{n}\sum_{i=1}^{n} W_q(\boldsymbol{x}_i) \cdot P(A|\boldsymbol{x}_i), \quad \boldsymbol{x}_i \sim q(\boldsymbol{x}),
\end{equation}
where $W_q(\boldsymbol{x})$ denotes the weight of each testing episode, i.e., likelihood ratio, ensuring the unbiasedness of testing results. It can be calculated by
\begin{equation}
W_q(\boldsymbol{x}) = \frac{P(\boldsymbol{x})}{q(\boldsymbol{x})} = \prod_{k=0}^{T} \left[\frac{P(\mathbf{u}(k)|\mathbf{s}(k))}{q(\mathbf{u}(k)|\mathbf{s}(k))}\right],
\end{equation}
where $q(\boldsymbol{u}(k)|\boldsymbol{s}(k))$ denotes BVs behavior distribution in the NADE.

To develop the NADE, we applied methods proposed in Ref.~\cite{feng2021intelligent}. The core idea is to identify the critical BV, referred to as the Principal Other Vehicle (POV), and adjust its behavior to challenge the AV. To identify the POV, we define the criticality value for the j-th BV at the k-th timestep as
\begin{equation}
C_j(\mathbf{s}(k)) = \sum_{\mathbf{u}_j(k)} V(\mathbf{u}_j(k)|\mathbf{s}(k)),
\end{equation}
where $\mathbf{u}_j(k)$ denotes the j-th BV's action, and $V(\mathbf{u}_j(k)|\mathbf{s}(k))$ denotes the maneuver criticality of the action, calculated as
\begin{equation}
V(\mathbf{u}_j(k)|\mathbf{s}(k)) = P(\mathbf{u}_j(k)|\mathbf{s}(k)) \cdot P(A_j|\mathbf{s}(k), \mathbf{u}_j(k)),
\end{equation}
where $A_j$ denotes the crash between the j-th BV and AV. The maneuver criticality value is the product of the BV's action exposure frequency $P(\mathbf{u}_j(k)|\mathbf{s}(k))$, which comes from the NDE model, and the corresponding maneuver challenge $P(A_j|\mathbf{s}(k),\mathbf{u}_j(k))$, which estimates the likelihood of a crash based on calibrated human driving behavior as a surrogate for the AV.
At each timestep, if at least one BV has a criticality value greater than zero, we define it as a critical moment, and the BV with the highest criticality is identified as the POV. To accelerate testing, all other BVs will maintain naturalistic behavior while the POV's behavior is intelligently modified. The importance function $q(\mathbf{u}(k)|\mathbf{s}(k))$ is developed as
\begin{equation}
q(\mathbf{u}(k)|\mathbf{s}(k)) = q(\mathbf{u}_j(k)|\mathbf{s}(k)) \cdot \prod_{m=1, m\neq j}^{M} P(\mathbf{u}_m(k)|\mathbf{s}(k)),
\end{equation}
where M represents the number of BVs. Specifically, the POV's adversarial maneuver probability is amplified by constructing its new behavior distribution as a weighted average of the naturalistic and normalized criticality distributions
\begin{equation}
q(\mathbf{u}_j(k)|\mathbf{s}(k)) = \epsilon \cdot \frac{V(\mathbf{u}_j(k)|\mathbf{s}(k))}{C_j(\mathbf{s}(k))}.
\end{equation}
Here, $\epsilon$ balances the exploration and exploitation of adversarial maneuvers, mitigating the impact of criticality approximation errors introduced by the surrogate model. In this study, $\epsilon$ is set to 3000, in line with the configuration in Ref.~\cite{feng2021intelligent}. Theoretical analysis in Ref.~\cite{feng2021intelligent} has also demonstrated that testing AVs using the proposed NADE framework is both efficient and unbiased, offering a solid theoretical foundation for the methodology.

\section{Behavioral Safety Evaluation}
\label{sec: evaluation}

\begin{figure*}[t]
  \centering
  \includegraphics[width=\textwidth]{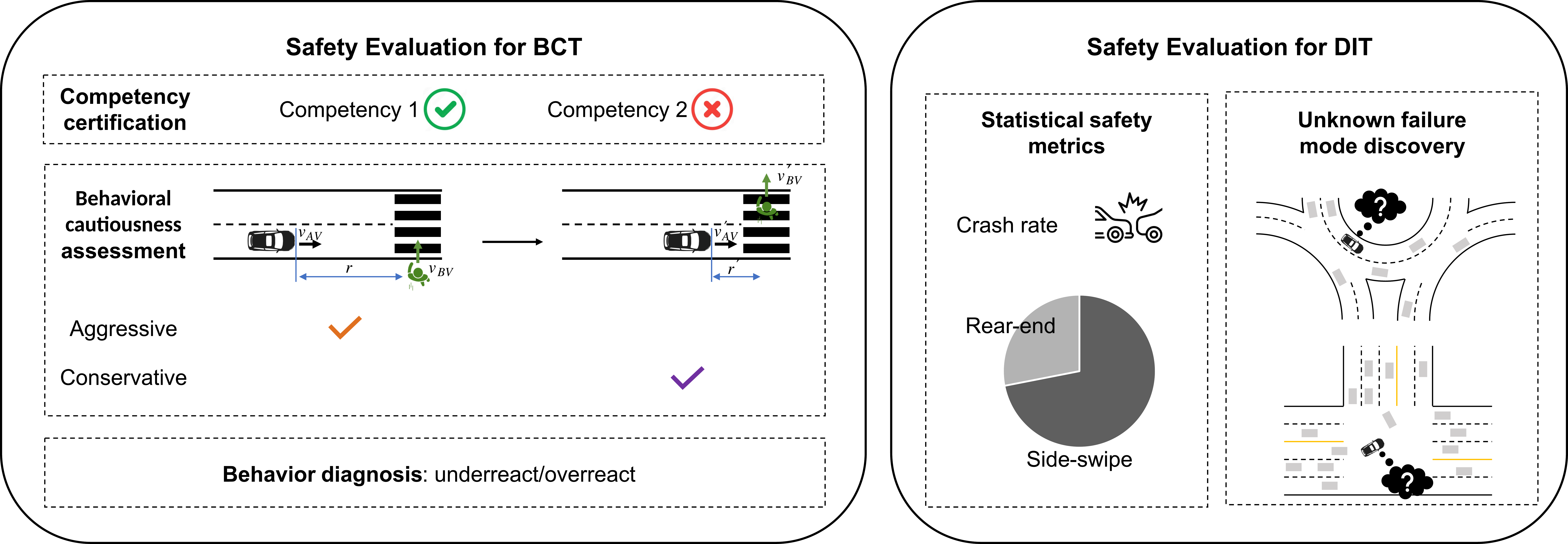}
  \caption{Behavioral safety evaluation frameworks}
  \label{fig: evaluation}
\end{figure*}

\subsection{Safety evaluation for BCT}
The BCT evaluates whether an AV possesses the basic behavioral competencies required for safe operation within its ODD.
It includes three aspects, as illustrated by the left box in Fig.~\ref{fig: evaluation}.
\begin{itemize}
    \item \textbf{Competency certification}. The primary output of the BCT is a formal determination of whether the AV possesses each behavioral competency, that is, whether the AV successfully avoids collision across all sampled test cases within the scenario.
    If the AV passes all test cases, it is certified to possess the corresponding behavioral competency at a quantifiable confidence level $\text{Conf}(n)$, as defined in Section~\ref{subsec: bct nade}.

    \item \textbf{Behavioral cautiousness assessment}. Beyond the binary certification outcome, the evaluation examines the AV's safety margins across the phases of a conflict event. Insufficient margins are problematic even when no collision occurs. Late braking for a pedestrian or planned near misses erode other road users' confidence and may cause them to hesitate. Frequent near misses also undermine public acceptance of AVs and hinder their adoption. Conversely, excessively wide margins are also undesirable. Beyond reducing traffic efficiency, overly conservative behavior, such as braking harder or earlier than surrounding drivers expect, may induce secondary collisions, particularly rear-end crashes into the AV. Two phases are generally considered, which may be further subdivided for specific competencies (e.g., acceleration, cruising, and deceleration phases in car following).
    \begin{itemize}
        \item \textit{Pre-encroachment phase}. Both road users are approaching the conflict zone. Metrics applicable to this phase capture the perceived imminence of a potential collision as well as the resulting safety margins once the conflict is resolved, with minimum Time-to-Collision (TTC), stopping clearance distance, and lateral passing distance serving as representative examples.
        \item \textit{Post-encroachment phase}. One road user has fully cleared the conflict zone and the conflict is unambiguously resolved. Metrics in this phase provide a retrospective characterization of how closely a collision was avoided, of which Post Encroachment Time (PET) is a widely adopted example.
    \end{itemize}

    \item \textbf{Behavioral diagnosis}. To provide actionable insight into the observed safety performance, the evaluation analyzes the AV's fine-grained behavioral responses, including but not limited to braking timing and deceleration profile. This diagnostic analysis identifies whether the AV underreacts, failing to respond adequately to the hazard presented, or overreacts, responding disproportionately relative to the actual risk level, thereby supporting targeted improvement of the AV's decision-making and control strategies.
\end{itemize}

\subsection{Safety evaluation for DIT}

The DIT assesses the AV's statistical safety performance within its ODD, going beyond predefined scenarios to uncover previously unknown failure modes. The evaluation encompasses two aspects, as illustrated by the right box in Fig.~\ref{fig: evaluation}.

\begin{itemize}
    \item \textbf{Statistical safety metrics}. The most fundamental output of the DIT is the crash rate $P(A)$, estimated using the importance sampling framework described in Section~\ref{subsec: dit nade}. This provides an unbiased measure of the frequency at which the AV encounters safety-critical events during naturalistic traffic interaction. Beyond the overall crash rate, the distribution of crash types and crash severity further enriches the evaluation, offering deeper insight into the nature and consequences of AV failures. Together, these statistical metrics characterize the AV's safety performance and provide the empirical basis for assessing its readiness for large-scale deployment.
    \item \textbf{Unknown failure mode discovery}. Beyond statistical measurement, a distinctive objective of the DIT is to identify previously unknown failure modes that may not be anticipated by predefined scenario-based testing. By exposing the AV to a broad range of naturalistic and adversarial driving conditions within the NADE, the DIT enables the emergence and characterization of unsafe events that arise unexpectedly from AV-environment interactions. Systematically analyzing these failure modes provides critical guidance for AV development and refinement, and is of particular importance before large-scale deployment, where unforeseen risks may carry significant societal consequences.
\end{itemize}

\section{Conclusion}

In this study, we present a systematic and standardized third-party testing framework for assessing AV behavioral safety, operating under a black-box approach that requires no prior knowledge of the AV’s internal workings.
The framework aims to create a comprehensive blueprint that unifies testing objectives, procedures, and measurements with executable methodologies that can be refined and expanded to support large-scale AV deployment.
This evaluation framework addresses both minimum performance requirements, such as behavioral competency, and comprehensive performance assessment, such as the occurrence rate of measurable safety-related events, like crash rates, to comprehensively evaluate AV behavior performance.
The testing scenarios and environments are designed to be both naturalistic and adversarial, substantially improving testing efficiency without sacrificing unbiasedness.

Part II of this study~\cite{Liu2026partII} complements this methodological foundation with a thorough simulation study and a field test conducted on an open-source Level 4 Automated Driving System, demonstrating the practical applicability of the proposed framework.

\section*{Acknowledgments}
This research was partially funded by the U.S. Department of Transportation (USDOT) Automated Driving System Demonstration Grant (\#693JJ319NF00001), USDOT Region 5 University Transportation Center: Center for Connected and Automated Transportation (CCAT) of the University of Michigan (\#69A3551747105), and the National Science Foundation (CMMI \#2223517). Any opinions, findings, conclusions, or recommendations expressed in this material are those of the authors and do not necessarily reflect the official policy or position of the U.S. government. 

We thank the late Prof. Huei Peng of the University of Michigan and Dr. Tony Geara of the City of Detroit for insightful discussions.

\bibliographystyle{IEEEtran}
\bibliography{ref}

\vfill

\end{document}